\documentclass[letterpaper]{article} % DO NOT CHANGE THIS
\usepackage{aaai25}  % DO NOT CHANGE THIS
\usepackage{times}  % DO NOT CHANGE THIS
\usepackage{helvet}  % DO NOT CHANGE THIS
\usepackage{courier}  % DO NOT CHANGE THIS
\usepackage[hyphens]{url}  % DO NOT CHANGE THIS
\usepackage{graphicx} % DO NOT CHANGE THIS
\usepackage{natbib}  % DO NOT CHANGE THIS AND DO NOT ADD ANY OPTIONS TO IT
\usepackage{caption} % DO NOT CHANGE THIS AND DO NOT ADD ANY OPTIONS TO IT
\usepackage{algorithm}
\usepackage{algorithmic}
\usepackage{amsmath}
\usepackage{multirow}
\usepackage{array}
\usepackage{booktabs}
\usepackage{tabularx}
\usepackage{subcaption}
\usepackage{makecell}
\usepackage{amssymb}
\usepackage{cleveref}
\usepackage{newfloat}
\usepackage{listings}
\DeclareCaptionStyle{ruled}{labelfont=normalfont,labelsep=colon,strut=off} % DO NOT CHANGE THIS
\floatstyle{ruled}
\newfloat{listing}{tb}{lst}{}
\floatname{listing}{Listing}
\title{InpDiffusion: Image Inpainting Localization via Conditional Diffusion Models}
\author{
    Kai Wang,
    Shaozhang Niu\thanks{Corresponding author.},
    Qixian Hao,
    Jiwei Zhang
}
\affiliations{
     Beijing Key Lab of Intelligent Telecommunication Software and Multimedia, Beijing University of Posts and Telecommunications, Beijing 100876, China\\
    {\rm \{kaiwang, szniu, qixianhao123, jwzhang666\}@bupt.edu.cn} 
}

\usepackage{bibentry}
\begin{document}

\maketitle
\begin{abstract}
As artificial intelligence advances rapidly, particularly with the advent of GANs and diffusion models, the accuracy of Image Inpainting Localization (IIL) has become increasingly challenging. Current IIL methods face two main challenges: a tendency towards overconfidence, leading to incorrect predictions; and difficulty in detecting subtle tampering boundaries in inpainted images. In response, we propose a new paradigm that treats IIL as a conditional mask generation task utilizing diffusion models. Our method, InpDiffusion, utilizes the denoising process enhanced by the integration of image semantic conditions to progressively refine predictions. During denoising, we employ edge conditions and introduce a novel edge supervision strategy to enhance the model's perception of edge details in inpainted objects. Balancing the diffusion model's stochastic sampling with edge supervision of tampered image regions mitigates the risk of incorrect predictions from overconfidence and prevents the loss of subtle boundaries that can result from overly stochastic processes. Furthermore, we propose an innovative Dual-stream Multi-scale Feature Extractor (DMFE) for extracting multi-scale features, enhancing feature representation by considering both semantic and edge conditions of the inpainted images. Extensive experiments across challenging datasets demonstrate that the InpDiffusion significantly outperforms existing state-of-the-art methods in IIL tasks, while also showcasing excellent generalization capabilities and robustness.
\end{abstract}

% Uncomment the following to link to your code, datasets, an extended version or similar.
%
% \begin{links}
%     \link{Code}{https://aaai.org/example/code}
%     \link{Datasets}{https://aaai.org/example/datasets}
%     \link{Extended version}{https://aaai.org/example/extended-version}
% \end{links}

\section{Introduction}

Image inpainting, essential for digital restoration and other applications, has reached unprecedented realism through advancements in deep learning techniques \cite{rombach2022high,zeng2022aggregated,feng2023hierarchical,park2024localization}. This sophistication has outpaced traditional tampering detection methods, which depend on visible signs. The field of image forensics now faces the new challenge of detecting subtle, concealed manipulations through the emerging task of Image Inpainting Localization (IIL).

\begin{figure}
    \centering
    \includegraphics[width=1\linewidth]{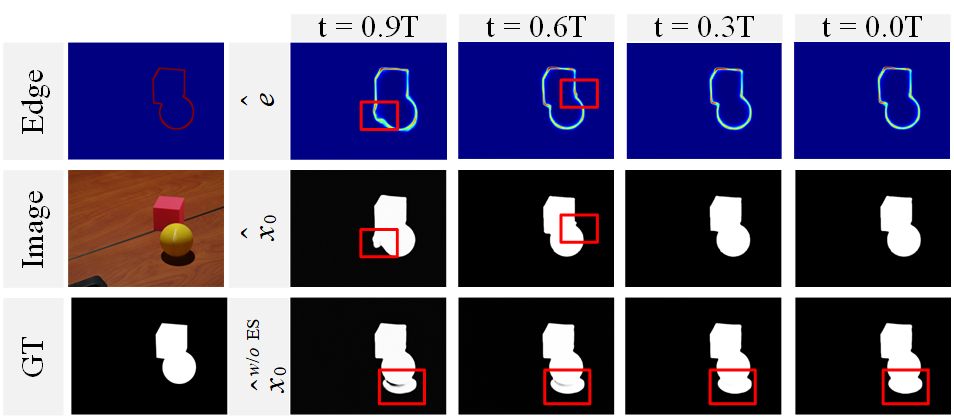}
    \caption{Visualization results of tampered edges and region predictions captured by InpDiffusion at different sampling stages. \(\hat x_0\) and \(\hat e\) represent the predicted inpainted objects and their edges at different sampling stages, respectively. \(\hat x_0^{w/o{\rm{ ES}}}\) are the predictions without edge supervision strategy
.}
    \label{fig:show}
\end{figure}

Researchers have responded to this challenge by proposing various strategies, focusing primarily on three design paradigms: 1) Multi-stream frameworks \cite{dong2022mvss,niloy2023cfl} that leverage multiple input streams to learn comprehensive representations; 2) Multi-stage \cite{liu2022pscc,guo2023hierarchical} feature fusion frameworks that integrate features extracted at different levels of the backbone network; 3) Branch frameworks \cite{xu2023up,shi2023transformer} that implement a single-input, multiple-output architecture, including tasks like word segmentation and auxiliary branches. These techniques are built upon a semantic segmentation foundation, employing learning-based backbones for feature extraction and decoder heads to generate segmentation masks. However, without a sophisticated design, this approach is prone to overconfidence in predictions, leading to incorrect assessments, especially when dealing with increasingly sophisticated inpainting images \cite{barglazan2024image}. The high invisibility of inpainting edges further complicates the task, as it blurs the distinction between the inpainted areas and the original image background, increasing the risk of misidentification.

To address the unique challenges of IIL, we develop a new framework called InpDiffusion. This framework leverages the diffusion model paradigm, known for its strong generative capabilities and sensitivity to conditions. Unlike traditional IIL models, InpDiffusion's iterative noise reduction mechanism simplifies the refinement process. Moreover, the diffusion model's stochastic sampling process introduces variability in predictions, mitigating the risk of overconfidence in the model's output. However, the direct application of diffusion models to IIL reveals certain limitations, such as limited discriminating power and excessive stochasticity in sampling, along with insufficient mask refinement. To address these limitations, we extract discriminative semantic and edge features from the inpainted images to guide the downstream denoising process. In addition, we introduce a novel edge supervision strategy within the denoising network to enhance InpDiffusion's ability to perceive subtle edges of locally inpainted objects. InpDiffusion generates both the denoised mask and the denoised edge of the inpainted object at each denoising step. By balancing the loss weights of these dual predictions, we aim to constrain the network and reduce detail loss caused by excessive randomness during each sampling step. More specifically, InpDiffusion integrates an Adaptive Conditional Network (ACN), which simultaneously extracts semantic and edge features from images. The Dual-stream Multi-scale Feature Extractor (DMFE) we designed is the core component of the ACN, enabling the adaptive extraction of multi-scale features to enhance the representation of both semantic and edge features. Then, the extracted semantic feature serves as a guiding clue for the denoising process, while the extracted edge feature is used for edge supervision, thereby strengthening the network's ability to capture fine details and reducing the randomness of each sample. This innovative dual approach is pivotal for identifying objects with concealed tampering traces, leading to more reliable and consistent results. As shown in \Cref{fig:show}, when edge supervision is absent, the model treats a small shaded area below the ball as a tampering area due to excessive random sampling. In contrast, the model initially produces rough edges of the tampering region, but with the progress of sampling and continuous constraints of edge features, the model finally gets a refined prediction. InpDiffusion offers several advantages over previous IIL methods: 1) Multiple random sampling of the diffusion model can avoide overconfident point estimation issues. 2) Efficient extraction of multi-scale features allows for a more comprehensive understanding of the intrinsic differences between inpainted objects and the image background. 3) The balance between image semantic supervision and image edge supervision significantly improves the handling of detail loss in inpainted areas.

Our main contributions can be summarized as follows:

\begin{itemize}
    \item We are the first to treat the IIL task as a mask generation paradigm and use a conditional diffusion framework to form predictions.
    \item We propose a novel and effective framework called InpDiffusion, which introduces an edge supervision strategy within the denoising network to constrain excessive stochasticity in sampling at each step. Additionally, we develop a Dual-stream Multi-scale Feature Extractor (DMFE) to extract key discriminative features, guiding the downstream denoising processes to produce more accurate and generalized results for IIL tasks.
    \item Comprehensive experiments confirms that InpDiffusion surpasses current benchmarks in IIL, while exhibiting remarkable adaptability across various scenarios.
\end{itemize}

\section{Related Work}
\subsection{Image Inpainting Localization}
Recently, models based on CNNs and Transformers have shown impressive results across various visual tasks, leading to an increase in the development of IIL. For example, IID-Net \cite{wu2021iid} uses a high-pass filter to extract features from the high-frequency domain. CAT-Net \cite{kwon2021cat} performs double compression detection on JPEG images to obtain an encoder with microscopic feature weights and a parallel combination of macroscopic feature weights to form a dual-stream network. MVSS-Net \cite{dong2022mvss} consists primarily of an edge supervision branch and a noise-sensitive branch, using the noise view, the boundary image, and the real image for feature learning. PSCC-Net \cite{liu2022pscc} employs a progressive mechanism to predict the operation masks on all scales. It uses a spatio-channel correlation module to guide feature extraction via space attention and channel attenuation. UE-Net \cite{ji2023uncertainty} is trained under dynamic supervision and produces estimated uncertainty maps to refine the detection results of manipulation, which significantly alleviates learning difficulties. TANet \cite{shi2023transformer} locates areas for image manipulation using an operator-induced approach, particularly when detecting and segmenting tampered areas that are cleverly embedded in normal images. EditGuard \cite{zhang2024editguard} embeds invisible watermarks to simultaneously achieve copyright protection and precise tamper localization, especially for AI-generated content editing methods. Despite progress, many IIL models remain constrained by overconfident point estimates and false segmentation. Moreover, their intricate structures require meticulous calibration to effectively address sophisticated tampering.

\subsection{Diffusion Models}
Diffusion Models, with its two-stage diffusion and reverse process, is a versatile framework for generating and manipulating data, including image generation \cite{dhariwal2021diffusion,nichol2021improved}, inpainting \cite{chung2022come,rombach2022high}, and editing \cite{avrahami2022blended,choi2021ilvr}. Its ability to learn from noise patterns has also made it a strong candidate for tasks such as super-resolution \cite{li2022srdiff,wang2021s3rp}, deblurring \cite{whang2022deblurring,lee2022progressive}, image segmentation \cite{baranchuk2021label,brempong2022denoising} and anomaly detection \cite{wolleb2022diffusion,wyatt2022anoddpm}, where noise analysis is key. Based on diffusion models, our paper introduces the innovative InpDiffusion model, which advances the field of IIL and significantly enhancing the capacity for detailed analysis within this domain.

\begin{figure*}
    \centering
    \includegraphics[width=1\linewidth]{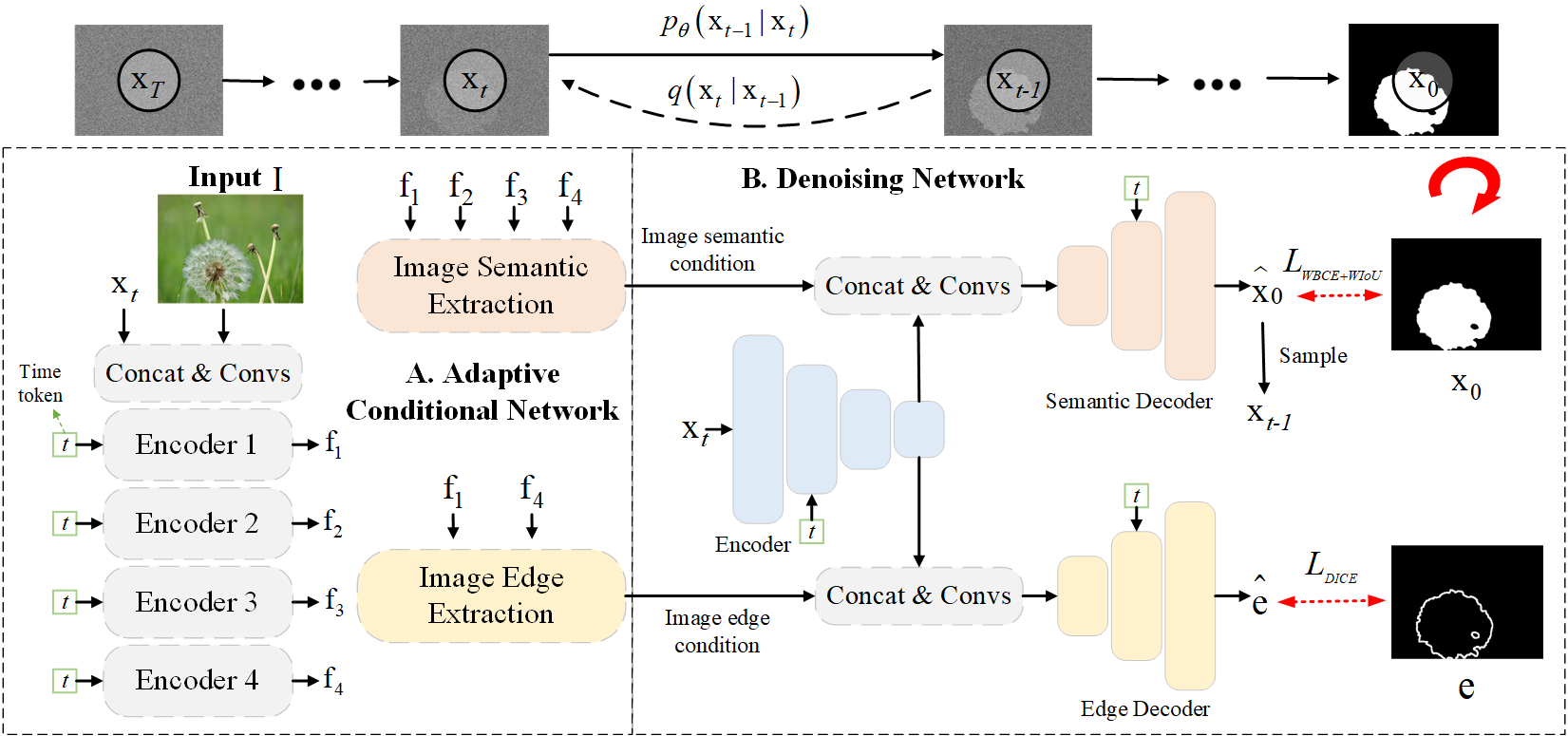}
    \caption{The framework of our InpDiffusion which includes an Adaptive Conditional Network (ACN), and a Denoising Network (DN). Instead of relying on the discriminative learning paradigm, our framework adopts a generative approach to guarantee reliability and generalizability.
}
    \label{fig:framwork}
\end{figure*}

\section{Method}

\subsubsection{Motivation.}The potential of generative models \cite{choi2021ilvr,ho2020denoising} for complex semantic understanding has been extensively explored, offering advanced pattern recognition and iterative refinement capabilities. These capabilities provide a fresh perspective for tackling the increasingly challenging task of IIL. Drawing inspiration from these insights, we are motivated to investigate the potential of diffusion models to further contribute to the IIL field.
\subsubsection{Overview.}
In this section, we introduce our InpDiffusion framework, which progressively generates predictions by using image semantic priors to condition each subsequent step while employing edge priors to constrain the sampling process at each step. As illustrated in Figure 2, our framework comprises two main components: an Adaptive Conditional Network (ACN) and a Denoising Network (DN). We begin by introducing the preliminary, followed by a detailed discussion of the architectures of ACN and DN. Finally, we introduce the loss function used in InpDiffusion.

\subsection{Preliminary}
Conventional IIL relies on a discriminative paradigm, classifying each pixel in an image $I \in \mathbb{R}^{\text{H} \times \text{W} \times 3}$ against an inpainted map $M \in \mathbb{R}^{\text{H} \times \text{W} \times 1}$. This approach uses a mapping function $F_\theta$, trained on labeled pairs $\{I_i, M_i\}_{i=1}^{N}$, to predict the presence ('1') or absence ('0') of inpainted areas. In contrast, our InpDiffusion framework is based on diffusion models, which consist of a forward diffusion process and a reverse denoising process. The forward diffusion process generates noisy masks from clean inputs by iteratively adding noise, following a Markov process. This process can be mathematically described as follows:
\begin{equation}
    q ( \mathbf x _ { t } | \mathbf  x _ { t-1} ) = \mathcal N ( \mathbf  x _ { t } ; \sqrt { 1 - \beta _ { t } } \mathbf  x _ { t-1 } , \beta _ { t }  \textbf I ) ,
\end{equation}
where ${\beta}_t$ denotes the variance parameters of the Gaussian noise added at each step $t$. As $t$ progresses from 1 to $T$, $\mathbf{x}_t$ evolves from the original inpainted map $\mathbf{x}_0$ through the process:
\begin{equation}
    q(\mathbf x_t|\mathbf x_0) = \mathcal N(\mathbf x_t;\sqrt{\bar\alpha_t} \mathbf x_0,(1-\bar \alpha_t) \textbf I),
\end{equation}
where $\bar \alpha_t=\prod_{i=1}^T\alpha_t, \alpha_t = 1-\beta_t$. 
During the training phase, we modulate the signal-to-noise ratio of the diffusion process by employing an SNR-based variance schedule \cite{hoogeboom2023simple}. For more detailed Settings and experiments on this aspect, please refer to \textit{Supplementary}.

In reverse denoising, diffusion models execute a sequence of transitions from ${\mathbf{x}_T} \rightarrow \mathbf{x}_{T-\Delta } \rightarrow \dots \rightarrow \mathbf{x}_{0}$, progressively refining initial noise $\mathbf{x}_T$ towards a refined inpainted map $\mathbf{x}_0$. Following \cite{song2020denoising}, we choose to train a network (InpDiffusion) $f_\theta(\mathbf x_t, \mathcal I, t)$ to directly estimate the denoised mask $\hat {\mathbf x}_0$ conditional on image $\mathcal I$. The network learns the reverse distribution:
\begin{equation}
    p(\mathbf x_{t-1}|\mathbf x_t):= \mathcal N(\mathbf x_{t-1} ; \mu_\theta(\mathbf x_t,t),\Sigma_\theta(\mathbf x_t,t)).
\end{equation}
Where, $\Sigma_\theta(\mathbf x_t, t)$ is set to $\sigma_t^2=\frac{1-\bar \alpha_{t-1}}{1-\bar \alpha_t}\beta_t$, and $\mu_\theta(\mathbf x_t,t)$ can be expressed as: 
\begin{equation}
    \mu_\theta(\mathbf x_t,t) = \frac{\sqrt{\alpha_t}(1 - \bar{\alpha}_{t-1})}{1 - \bar{\alpha}_t} \mathbf{x}_t + \frac{\sqrt{\bar{\alpha}_{t-1}}\beta_t}{1 - \bar{\alpha}_t} \hat{\mathbf{x}}_0.
\label{con:mu_theta}
\end{equation}
where $\hat{\mathbf{x}}_0$ is predicted by InpDiffusion.

\subsection{Adaptive Conditional Network (ACN)}

The primary function of the Adaptive Convolutional Network (ACN) is to adaptively extract semantic and edge features from inpainted images, allowing the downstream denoising network to recognize inpainted targets and their edges with greater accuracy. The entire ACN consists of two main components: \textbf{Hierarchical Feature Extraction} and \textbf{Image Semantic and Edge Extraction}.
\subsubsection{Hierarchical Feature Extraction.}
Hidden tampering traces in inpainted images pose significant challenges for feature extraction by the model, which leads to a decline in the mask decoder's performance. To address this, we input the coarsely predicted mask $\mathbf x_t$ from the previous step along with the inpainted image $\mathcal I$ into the ACN, using $\mathbf x_t$ as a guiding cue to focus the network on specific regions. Additionally, to enhance feature adaptability across denoising steps, we incorporate the time token (denoted as $t$) into the feature extraction process. This enables the ACN to autonomously adjust the conditional features according to the temporal stage. As shown in \Cref{fig:framwork} A., we use the Pyramid Vision Transformer (PVT) \cite{wang2022pvt} as the backbone of the ACN to extract multi-scale features ($\left\{\rm{f}_i\right\}_{\rm{i}=1}^4$) from concatenated $\mathcal I$ and $\mathbf x_t$. Thanks to the Transformer's token-based design, we can treat time as a token and incorporate it into each layer of patch sequences obtained from the backbone, which are then fed into the corresponding Transformer encoders. Formally, \({{\mathbf{f}}_i}{\rm{  =  }}{{\rm{R}}^{ - 1}}{\rm{(FN(MA([t; }}{{\rm{f}}_i}{\rm{]))).}}\) Here, $\left[ { \cdot  ;  \cdot } \right]$ refers to the concatenation operation, and ${\rm{R}^{ - 1}}$ converts tokens to feature maps. MA represents multi-head self-attention. FN denotes the feedforward neural network. $\left\{\rm{f}_i\right\}_{\rm{i}=1}^4$ are the hierarchical features we end up with.

\begin{figure}
    \centering
    \includegraphics[width=1\linewidth]{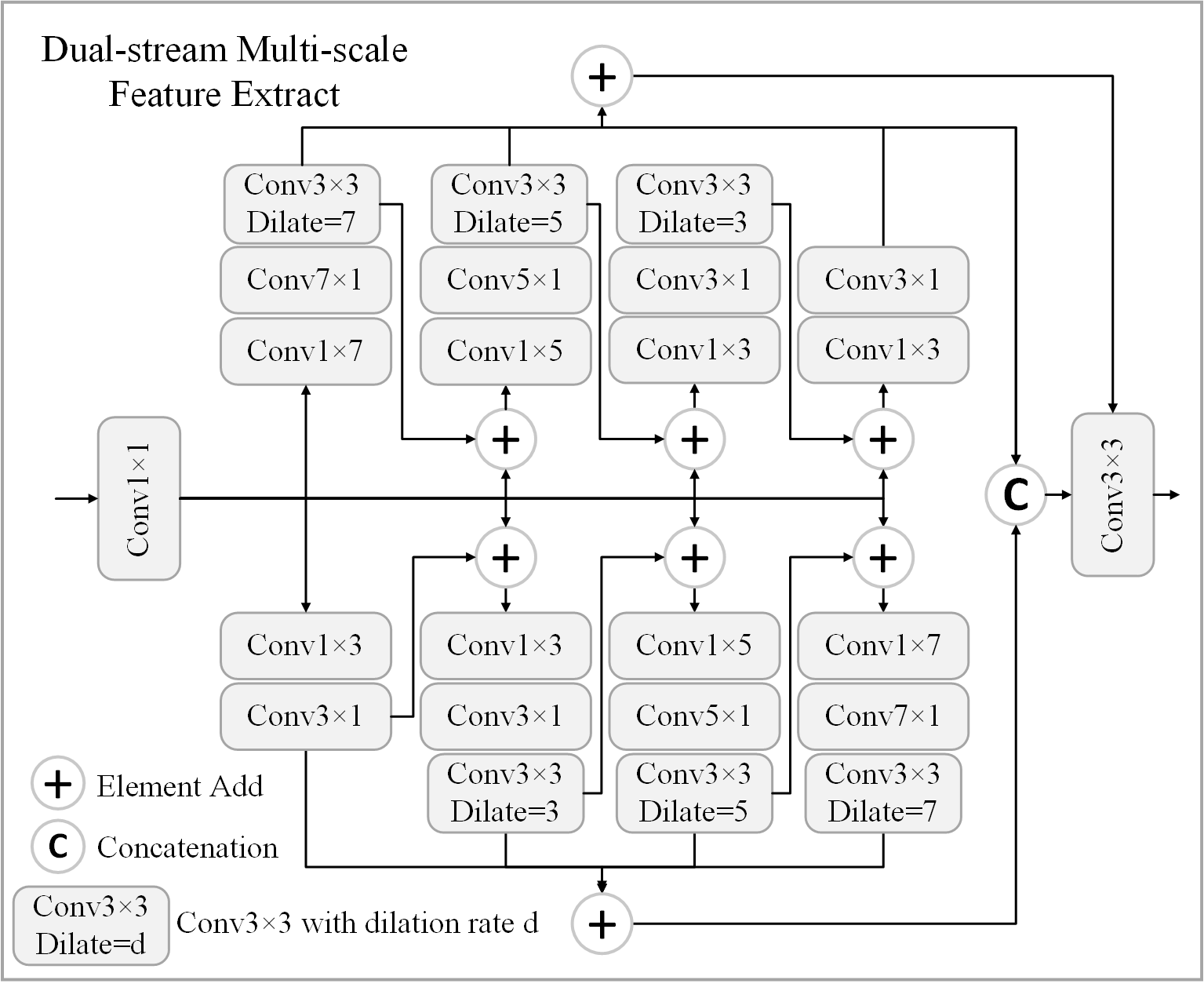}
    \caption{Illustration of Dual-stream Multi-scale Feature Extractor (DMFE) 
.}
    \label{fig:DMFE}
\end{figure}

\subsubsection{Image Semantic and Edge Extraction.}
\begin{figure}
    \centering
    \includegraphics[width=1\linewidth]{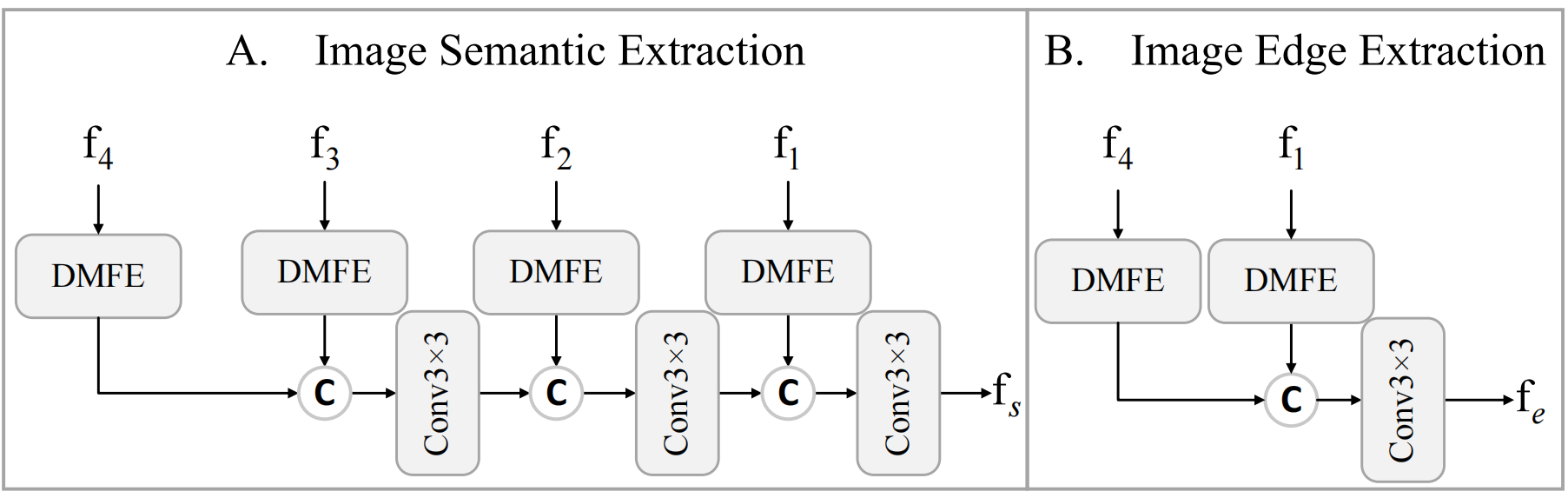}
    \caption{Illustration of Image Semantic and Edge Extraction 
.}
    \label{fig:CE}
\end{figure}

The extraction of key discriminative features is essential for denoising networks to accurately identify inpainted targets. To enhance this process, we propose the \textbf{Dual-stream Multi-scale Feature Extractor (DMFE)} module as a core component for adaptively extracting image semantic and edge conditions, enabling the capture of additional multi-scale features that the backbone network may not extract. The illustration of Image Semantic and Edge Extraction is shown in \Cref{fig:CE}. $\left\{\rm{f}_i\right\}_{\rm{i}=1}^4$ are used to extract semantic feature layer by layer based on DMFE. For the edge condition, we merge the detail-oriented low-level feature $\rm{f}_1$ with the spatially informative high-level feature $\rm{f}_4$, to accurately represent the edge feature associated with the inpainted object.

\normalfont{\normalsize\bfseries \textbullet\ \ Dual-stream Multi-scale Feature Extractor:}
Given that the PVT-based backbone network may not extract abundant context information, we drew inspiration from the Inception module and Res2Net module \cite{gao2019res2net} and subsequently developed a two-stream multi-scale feature extractor (DMFE). The DMFE consists of two streams. The first stream incrementally applies dilated convolution with a progressively increasing dilated convolution rate, thereby gradually widening the contextual understanding. In contrast, the second stream uses dilated convolution processing with a gradually decreasing rate, which first captures a broad context and then slowly narrows down on specific details. More specifically, we first reduce the number of channels of the input feature $f$ using a $1 \times 1$ convolution to facilitate subsequent processing. In the first stream, we utilize four branches to capture features at different scales, sequentially processed with a dilated convolution set with a dilation rate of $\left\{ {3,5,7} \right\}$ . Each branch is equipped with an asymmetric convolution that matches the size of the dilated convolution, reducing computational effort. The output of each branch is added to the input of the next branch. Concurrently, we use residual connections to enlarge the receptive field successively. The general form of an operation is defined as follows:
\begin{equation}
Dout_k\left\{ \begin{array}{l}
Co{v_{3 \times 1}}(Co{v_{1 \times 3}}(Cov(f))){\rm{   k = 1 }}\\
Co{v_s}(Cov(f) + Dout_{k - 1}){\rm{    k = 2,3,4}}
\end{array} \right.{\rm{ }}.
\end{equation}
where $k$ is the branch number, $Dout_k( \cdot )$ represents the output of the $kth$ branch in the first stream,  $ + $ refers to element addition operation. $Co{v_s}( \cdot )$ denotes the stacked convolutional layer mentioned above. $Co{v_{3 \times 1}}( \cdot )$ refers to $3 \times 1$ convolution operation,   $Co{v_{1 \times 3}}( \cdot )$ refers to $1 \times 3$ convolution operation. In contrast to the first stream, the second stream, in turn, processes the features using a dilated convolution set with a dilation rate of $\left\{ {7,5,3} \right\}$:
\begin{equation}
Uout_k\left\{ \begin{array}{l}
Co{v_s}(Cov(f)){\rm{   }}k{\rm{ = 1 }}\\
Co{v_s}(Cov(f) + Uout_{k - 1}){\rm{    }}k{\rm{ = 2,3}}\\
Co{v_{3 \times 1}}(Co{v_{1 \times 3}}(Cov(f) + Uout_{k - 1})){\rm{    }}k{\rm{ = 4}}
\end{array} \right.{\rm{ }}.
\end{equation}
where $Uout_k( \cdot )$ denotes the output of the $kth$ branch in the second stream, $Cov( \cdot )$ refers to  $1 \times 1$ convolution operation. After that, we add the output of the four branches for each stream to get the result. Finally, we concatenate the results of the two streams followed by a $Co{v_{3 \times 3}}$ to obtain the output feature, which is computed as:
\begin{equation}
{f^m} = Co{v_{3 \times 3}}(\left[ {(Add_{k = 1}^4(Dout_k),Add_{k = 1}^4(Uout_k)} \right]){\rm{ }}{\rm{.}}
\end{equation}
where $Add_{k = 1}^4( \cdot )$ refers to the element addition of all four branches.

\subsection{Denoising Network (DN)}
Traditionally, Denoising Networks (DN) aim to decode the denoised mask predictions $\hat{\mathbf x}_0$ and $\mathbf x_{t-1}$ based on the diffusion paradigm. The random sampling in the denoising network may result in insufficient model sensitivity to the edges of concealed inpainted objects. In contrast, our denoising network simultaneously predicts the edge $\hat{\mathbf e}$ of inpainted objects while generating denoised masks. Through edge loss supervision, we can balance the denoising process of the noised mask, thereby reducing detail loss caused by excessive randomness at each sampling step. As shown in \Cref{fig:framwork} B., our DN comprises an encoder and two decoders. The encoder extracts the positional information of inpainted objects from the noisy mask. One decoder, informed by the image semantic conditions extracted by the ACN, generates the final denoised mask predictions $\hat{\mathbf x}_0$ and $\mathbf x_{t-1}$. The other decoder, utilizing the image edge conditions extracted by the ACN, produces the final denoised edge predictions $\hat{\mathbf e}$. Notably, we employ adaptive group normalization \cite{dhariwal2021diffusion} in both the encoder and decoder to incorporate the time step information $t$ into the convolutional layers, making our DN sensitive to time step variations. Due to the iterative nature of the denoising process, our encoder and decoders employ a simple U-shaped structure, which has proven effective in achieving satisfactory results.
\subsection{Loss Function}
To ensure the denoising network generates accurate results, we measure the loss between the conditional denoising mask output $\hat{\mathbf x}_0$ and the ground truth  $\mathbf x_0$, as well as between the conditional denoising edge output $\hat{\mathbf e}$ and the ground truth $\mathbf e$. These losses guide the optimization of IpnDiffusion. In our work, we introduce the Weighted Binary Cross-Entropy (WBCE) and Weighted Intersection over Union (WIoU) loss functions \cite{wei2020f3net} for supervising the denoising result $\hat{\mathbf x}_0$ of each iteration of the network. For edge supervision, the Dice loss function \cite{xie2020segmenting} is used to handle the imbalance between positive and negative samples. The overall loss function for InpDiffusion is formulated as follows:

\begin{equation}
L_{\text{total}} = \lambda L_{\text{WBCE + WIoU}}\left( \hat{\mathbf{x}}_0, \mathbf{x}_0 \right) \\
+ \mu L_{\text{DICE}}\left( \hat{e}, e \right).
\end{equation}
where, $L_{\text{WBCE + WIoU}}( \cdot )$ refers to the combination of WIoU and WBCE loss functions, $ L_{\text{DICE}}( \cdot )$ refers to Dice loss function, $\lambda$ and $\mu $ are adjustment parameters. We achieve optimal denoising by adjusting the ratio of $\lambda$ to $\mu $. 
\begin{table}[tb]
    \centering
    \renewcommand{\arraystretch}{1.1}
    \setlength{\tabcolsep}{3pt}
    \small
    \begin{tabularx}{\linewidth}{>{\arraybackslash}p{4cm} *{4}{>{\centering\arraybackslash}X}}
        \Xhline{2\arrayrulewidth}
        \multicolumn{1}{l}{\multirow{2}{*}{\textbf{Baseline Models}}} & \multicolumn{4}{c}{\textbf{Inpaint 32K}} \\
        \cmidrule(lr){2-5}
        & \textbf{TM} & \textbf{CNN} & \textbf{GAN} & \textbf{DM} \\
        \hline
        MantraNet\cite{wu2019mantra} & 67.3 & 77.1 & 76.5 & 66.2 \\
        MVSSNet\cite{dong2022mvss} & 73.6 & 84.3 & 79.4 & 70.9 \\
        TANet\cite{shi2023transformer} & 85.1 & 87.2 & 82.7 & 73.1 \\
        MFINet\cite{ren2023mfi} & 84.7 & 91.1 & 87.9 & 75.3 \\
        CFLNet\cite{niloy2023cfl} & 79.0 & 88.5 & 83.4 & 72.6 \\
        ECNet\cite{hao2024ec} & 86.4 & 90.5 & 88.3 & 78.1 \\
        \hline
        Ours-w/o ES & 91.6 & 93.3 & 93.7 & 82.4 \\ 
        \multicolumn{1}{l}{Ours} & \textbf{93.7} & \textbf{96.2} & \textbf{95.6} & \textbf{84.3} \\
        \Xhline{2\arrayrulewidth}
    \end{tabularx}
    \normalsize
    \caption{Quantitative results of our method and other baseline methods across different inpainting types in the Inpaint32K dataset. TM, CNN, GAN, and DM represent four different inpainting types: Traditional Methods-Based, Convolutional Neural Network-Based, Generative Adversarial Network-Based, and Diffusion Model-Based, respectively. ES stands for the edge supervision strategy proposed in this paper. Results are in \% AUC.}
    \label{tab:quantitative-results}
\end{table}

\section{Experiments}
\begin{figure*}
    \centering
    \includegraphics[width=1\linewidth]{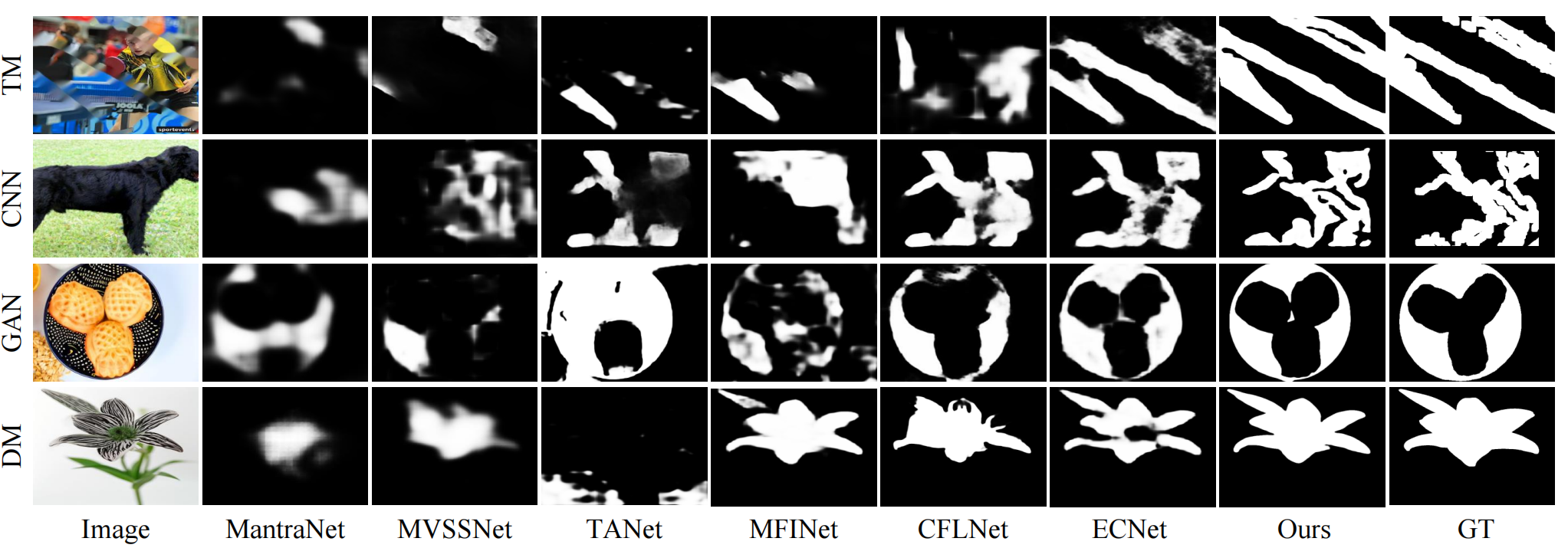}
    \caption{Visual comparisons with recent SOTA models in challenging scenarios with different inpainting techniques.
}
    \label{fig:comparison}
\end{figure*}

\begin{table*}[ht]
    \centering
    \renewcommand{\arraystretch}{1.1}
    \setlength{\tabcolsep}{4pt}
    \small
    \begin{tabularx}{\linewidth}{>{\arraybackslash}l *{1}{>{\centering\arraybackslash}X} *{10}{>{\centering\arraybackslash}X}}
        \Xhline{2\arrayrulewidth}
        \multicolumn{2}{l}{\multirow{2}{*}{\textbf{Baseline Models}}} & \multicolumn{10}{c}{\textbf{DID}} \\
        \cmidrule(lr){3-12}
        & & \textbf{GC} & \textbf{CA} & \textbf{SH} & \textbf{EC} & \textbf{LB} & \textbf{RN} & \textbf{NS} & \textbf{LR} & \textbf{PM} & \textbf{SG} \\
        \hline
        MantraNet \cite{wu2019mantra} & & 53.3 & 54.2 & 65.7 & 70.3 & 60.9 & 50.2 & 61.5 & 58.6 & 59.3 & 62.2 \\
        MVSSNet \cite{dong2022mvss} & & 59.5 & 72.9 & 81.1 & 86.0 & 75.4 & 68.3 & \textbf{85.5} & 74.2 & 79.9 & 91.8 \\
        TANet \cite{shi2023transformer} & & 73.6 & 90.2 & 93.1 & 90.6 & 94.3 & 90.8 & 83.6 & 84.5 & 90.7 & 93.4 \\
        MFINet \cite{ren2023mfi} & & 72.5 & 88.3 & 90.1 & 93.3 & 92.7 & 88.3 & 79.2 & 85.7 & 87.6 & 94.1 \\
        CFLNet \cite{niloy2023cfl} & & 58.8 & 84.6 & 84.3 & 86.8 & 86.4 & 65.6 & 84.3 & 78.2 & 82.5 & 93.3 \\
        ECNet \cite{hao2024ec} & & 69.5 & 87.3 & 91.1 & 91.8 & 91.6 & 86.3 & 80.1 & 77.7 & 85.4 & 92.3 \\
        \hline
        Ours-w/o ES & & 77.3 & 92.1 & 93.3 & 94.7 & 95.2 & 90.4 & 82.6 & 87.5 & 93.6 & 94.4 \\
        \multicolumn{2}{l}{Ours} & \textbf{79.4} & \textbf{94.5} & \textbf{96.1} & \textbf{97.2} & \textbf{98.0} & \textbf{93.2} & 84.9 & \textbf{91.3} & \textbf{96.7} & \textbf{98.3} \\
        \Xhline{2\arrayrulewidth}
    \end{tabularx}
    \normalsize
    \caption{Quantitative comparison of ten different inpainting techniques on the DID dataset under pre-training settings. Results are presented as \% AUC.}
    \label{tab:quantitative-results2}
\end{table*}
In this section, we describe the experiments conducted on five different datasets to evaluate the effectiveness of InpDiffusion. Our experiments are primarily conducted on a large dataset called Inpaint32K \cite{Inpaint32K}, which contains a total of 32,000 inpainted images. This dataset is divided into four distinct inpainting types, with each type comprising 8,000 images: Traditional Methods-Based, CNN-Based, GAN-Based, and Diffusion Model-Based. Other datasets include an inpainting datasets (DID \cite{wu2021iid}), an AIGC dataset (AutoSplice \cite{jia2023autosplice}), a real-life dataset (IMD \cite{novozamsky2020imd2020}), and a traditional benchmark dataset (Nist \cite{nist_nimble_2016}). We compare our model with six baseline models. MantraNet \cite{wu2019mantra}, MVSSNet \cite{dong2022mvss}, TANet \cite{shi2023transformer}, MFINet \cite{ren2023mfi}, CFLNet \cite{niloy2023cfl}, and ECNet \cite{hao2024ec}. More details about the datasets and baseline models are provided in \textit{Supplementary}. We conduct separate ablation and comparative experiments for each inpainting type in Inpaint32K dataset. Following this, we pre-train our model using 32K inpainted images from the Inpaint32K dataset and evaluate it on the DID dataset. Notably, the other baseline models are also re-trained using the Inpaint32K, even if some are pre-trained on larger datasets. To further demonstrate the generalization performance of our model, we fine-tune it on other datasets of varying types. It's important to note that DID is exclusively used as a test set, while the other datasets are divided into training and test sets with a 9:1 ratio. We use the pixel-level Area Under the Curve (AUC) score as the evaluation metric.
\subsection{Implementation Details}
We implement our InpDiffusion based on PyTorch using a single NVIDIA A800 with 80GB memory for both training and inference. For efficient training, the model undergoes a total of 150 training epochs. For optimization, the AdamW~\cite{loshchilov2017decoupled} optimizer was utilized along with a batch size set to 32. To adjust the learning rate, we implemented the cosine strategy with an initial learning rate of 0.001. Notably, we set T = 10 for sampling and set SNR Shift to $-2\log(6)$. In the loss function, we set the ratio of $\lambda$ to $\mu$ as 7:3. The ablation of hyperparameters we put in \textit{Supplementary}.

\subsection{Experiment Results}
\subsubsection{Quantitative Evaluations.} \Cref{tab:quantitative-results} shows the quantitative results of our proposed InpDiffusion compared to six other state-of-the-art (SOTA) methods across four different inpainting types on the Inpaint32K dataset. Impressively, across all types of inpainting, InpDiffusion consistently outperformed other models in AUC score, demonstrating significant superiority even compared to the next best-performing model. Notably, when our proposed edge supervision strategy is integrated into the denoising network, the model's performance improves significantly across each inpainting type. This further highlights the critical role of edge guidance in accurately locating inpainted images with hidden tampering traces. \Cref{tab:quantitative-results2} outlines the detection performance of InpDiffusion and other SOTA models on the DID dataset across ten different inpainting techniques, following pre-training on the Inpaint32K dataset. InpDiffusion consistently achieved superior accuracy in nearly all local inpainting techniques. However, MVSSNet exhibited slightly better performance in detecting images inpainted with the NS technique. We attribute this advantage to MVSSNet's extra pre-training on a larger dataset of 60K images, which enabled it to capture additional key information relevant to IIL.

\subsubsection{Qualitative Evaluation.}
 \Cref{fig:comparison} illustrates a comparative visual analysis of InpDiffusion against six established IIL baselines. In this figure, each row illustrates a typical inpainting scenario, with each inpainted region showing complex topological structures and edges. Compared to the first row, which employs traditional methods, and the second row, which utilizes CNN techniques, the inpainted regions in both cases do not fully form a complete semantic structure, presenting a significant challenge for current IIL models. However, the proposed InpDiffusion model excels in mitigating overly confident missegmentations caused by the random sampling inherent in each denoising step. Furthermore, the model effectively delineates clear boundaries due to its iterative denoising paradigm and edge supervision strategy.
\subsubsection{Ablation Studies.}
We conduct an ablation study on the individual components of InpDiffusion, with detailed results provided in \Cref{tab:ablation-results}. The absence of the DMFE module led to a decline in model performance, resulting in a 2.9\% drop in AUC score, particularly on inpainted images based on CNN. This underscores the crucial role of DMFE in extracting discriminative image semantic and edge features. The inclusion of DMFE enhances the feature representation of locally inpainted regions, enabling the downstream denoising network to accurately identify and locate these areas. The lack of edge supervision also caused a decrease in InpDiffusion's detection performance across all types of locally inpainted images, highlighting the importance of edge supervision in complementing the denoising network's ability to generate precise prediction maps.

\subsubsection{Generalization Performance Evaluation.}
To further evaluate the generalization performance of InpDiffusion and verify its scalability, we fine-tuned InpDiffusion and other SOTA methods using three different datasets involving other types of tampering. The experimental results, as shown in \Cref{tab:comparison-results}, indicate that our model continues to outperform other methods in localization performance. In particular, on the challenging real-life dataset IMD, our approach achieved a 10.2\% higher AUC score than the second-best method, representing a significant improvement.

\begin{table}[tb]
    \centering
    \renewcommand{\arraystretch}{1.2}
    \setlength{\tabcolsep}{1pt}
    \small
    \begin{tabularx}{\linewidth}{>{\centering\arraybackslash}p{1cm} >{\centering\arraybackslash}p{1cm} *{4}{>{\centering\arraybackslash}X}}
        \Xhline{2\arrayrulewidth}
        \multicolumn{2}{c}{\textbf{Components}} & \multicolumn{4}{c}{\textbf{Inpaint 32K}} \\
        \cmidrule(lr){1-2} \cmidrule(lr){3-6}
        \textbf{DMFE} & \textbf{ES} & \textbf{TM} & \textbf{CNN} & \textbf{GAN} & \textbf{DM} \\
        \hline
        $\times$ & $\times$ & 88.4 & 92.6 & 91.3 & 80.1 \\
        $\times$ & $\checkmark$ & 91.9 & 94.0 & 93.5 & 82.8 \\
        $\checkmark$ & $\times$ & 91.6 & 93.3 & 93.7 & 82.4 \\
        $\checkmark$ & $\checkmark$ & \textbf{93.7} & \textbf{96.2} & \textbf{95.6} & \textbf{84.3} \\
        \Xhline{2\arrayrulewidth}
    \end{tabularx}
    \normalsize
    \caption{Ablation of model components on Inpaint32K datasets. Results are in \% AUC.}
    \label{tab:ablation-results}
\end{table}
\begin{table}[tb]
    \centering
    \renewcommand{\arraystretch}{1.2}
    \small
    \setlength{\tabcolsep}{1pt}
    \begin{tabularx}{\linewidth}{>{\arraybackslash}p{3.8cm} *{3}{>{\centering\arraybackslash}X}}
        \Xhline{2\arrayrulewidth}
        \textbf{Baseline Models} & \textbf{AutoSplice} & \textbf{IMD} & \textbf{Nist} \\
        \hline
        MantraNet\cite{wu2019mantra} & 88.2 & 59.5 & 75.7 \\
        MVSSNet\cite{dong2022mvss} & 93.0 & 68.2 & 94.2 \\
        TANet\cite{shi2023transformer} & 93.6 & 71.7 & 93.3 \\
        MFINet\cite{ren2023mfi} & 96.1 & 73.5 & 95.8 \\
        CFLNet\cite{niloy2023cfl} & 93.2 & 71.8 & 94.6 \\
        ECNet\cite{hao2024ec} & 95.4 & 75.3 & 94.2 \\
        Ours & \textbf{97.3} & \textbf{85.5} & \textbf{96.8} \\
        \Xhline{2\arrayrulewidth}
    \end{tabularx}
    \normalsize
    \caption{Quantitative comparison of the other three different kinds of datasets in the fine-tuning setting. Results are in \% AUC.}
    \label{tab:comparison-results}
\end{table}
\begin{figure}[tb]
    \centering
    \includegraphics[width=1\linewidth]{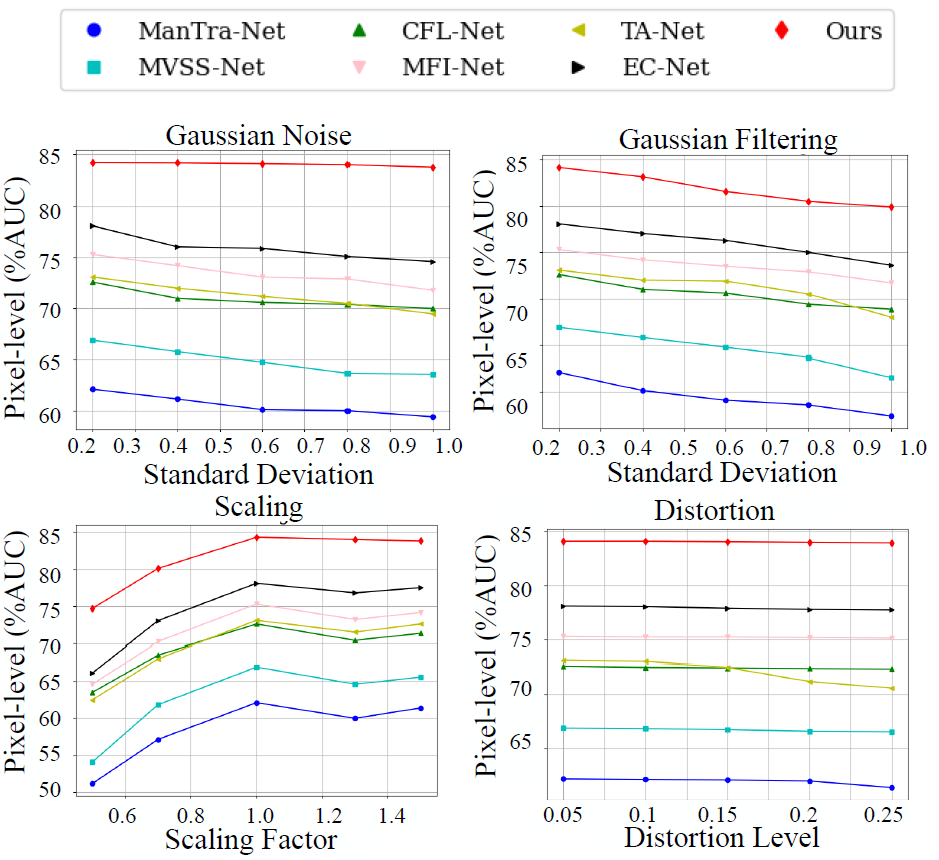}
    \caption{Robustness of seven methods against Gaussian Noise, Gaussian filtering, Scaling and Distortion. The detection is performed on the DM-based inpainted images in Inpaint32K. Our method achieves a substantial lead in IIL performance.}
    \label{fig:robt}
\end{figure}
\subsubsection{Robustness Evaluation.}
To verify the robustness of InpDiffusion, we conducted experimental evaluations on two types of image attack methods. The first type includes conventional image processing operations such as Gaussian noise and Gaussian filtering, which result in blurred image details and reduced contrast, creating fuzzy boundaries around the tampered areas and making it difficult for the model to accurately locate the tampered regions. The second type involves geometric transformation attacks, such as scaling and distortion, which alter the size, shape, and position of objects in the image, challenging the algorithm's ability to accurately identify the boundaries of the tampered areas. We tested our model on the relatively difficult-to-detect DM-based inpainted images in Inpaint32K. The experimental results are shown in \Cref{fig:robt}. Our method maintained high robustness against all types of image attacks, particularly against Gaussian noise and image distortion, where InpDiffusion was nearly unaffected by these attacks. We attribute this exceptional performance to InpDiffusion's use of joint supervision from the inpainted object's mask and edges to guide the step-by-step denoising process.

\section{Conclusion}
In this study, we propose a diffusion-based IIL model named InpDiffusion. InpDiffusion extracts semantic and edge features from inpainted images to guide the denoising process and progressively refine predictions. The introduction of edge supervision helps mitigate the risk of overconfident mispredictions and prevents the loss of subtle boundaries due to excessive randomness. We also introduce an innovative dual stream multiscale feature extractor to capture multi-scale features, enhancing feature representation. Extensive experiments demonstrate that InpDiffusion achieves SOTA performance, showcasing excellent generalization ability and robustness.

\section{Acknowledgments}
This work is supported by the National Natural Science Foundation of China(No.61370195) and the Joint Funds of the National Natural Science Foundation of China (No.U1536121).

\bibliography{aaai25}
\clearpage
\appendix
\twocolumn[{
    \section*{\Huge{Supplementary Material}}
    \vspace{10mm}
}]
In this supplementary material, we provide experimental details that were not included in our main paper due to space limitations.
\vspace{3mm}

% %%%%%%%%%%%%%%%%%%%%%%%%%%%%%%
% A. Additional Experimental Details
% %%%%%%%%%%%%%%%%%%%%%%%%%%%%%%

\section{Additional Experimental Details}
\label{sec:supp_experimental_details}
\setcounter{figure}{0}
\setcounter{table}{0}
\setcounter{algorithm}{0}
\setcounter{equation}{0}
\renewcommand{\thefigure}{A\arabic{figure}}
\renewcommand{\thetable}{A\arabic{table}}
\renewcommand{\thealgorithm}{A\arabic{algorithm}}
\renewcommand{\theequation}{A\arabic{equation}}

In this section, we provide a comprehensive description of implementation details, along with a range of supplementary experiments for deeper insights.

\subsection{Baselines}
The proposed method is compared with six deep learning-based methods for a fair comparison, which are listed as follows:
\begin{figure}[tb]
    \centering
    \includegraphics[width=1\linewidth]{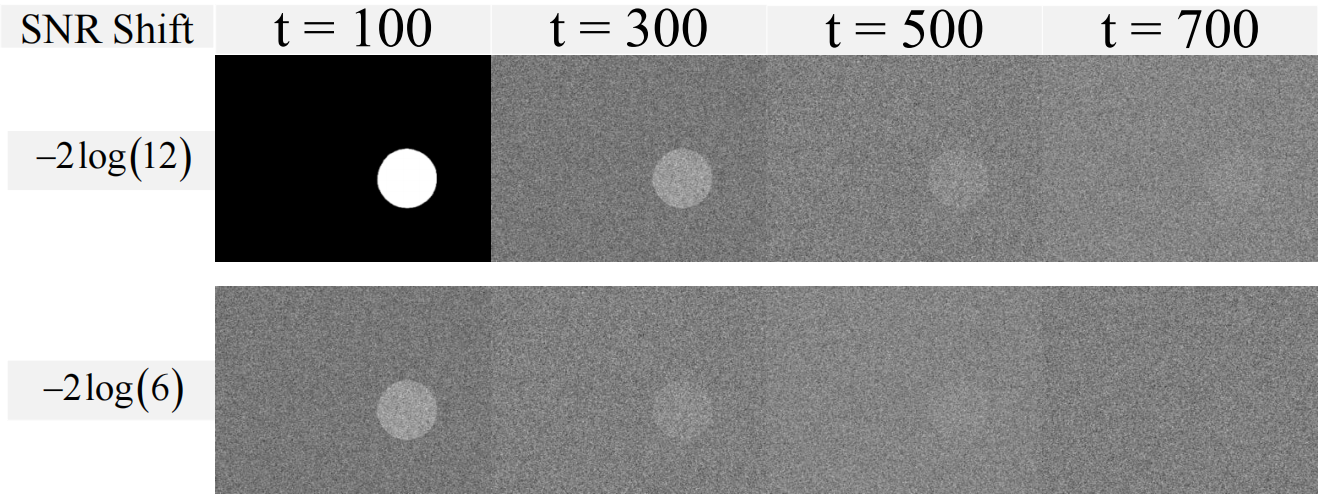}
    \caption{Visualization comparisons of noisy masks changing over time t under various SNR shift settings.}
    \label{fig:SNR shift}
\end{figure}

\begin{itemize}
    \item ManTraNet \cite{wu2019mantra}: ManTraNet uses a feature extractor to capture the manipulation traces and a local anomaly detection network to localize the manipulated regions.
    \item  MVSSNet \cite{dong2022mvss}: MVSSNet identify manipulated regions within images by employing multi-scale and multi-view analysis techniques.
    \item TANet \cite{shi2023transformer}: TANet is a Transformer-Auxiliary Neural Network that employs multi-scale transformers and operator inductions to accurately localize manipulated regions in images by detecting subtle inconsistencies and enhancing boundary details.
    \item MFINet \cite{ren2023mfi}: MFI-Net is an advanced multi-feature fusion identification network designed to effectively detect and localize artificially manipulated images by integrating detail extraction, multi-branch attention fusion, feature decoding, and detail enhancement modules.
    \item CFLNet \cite{niloy2023cfl}:CFL-Net is an innovative image forgery localization method that employs contrastive learning to distinguish between tampered and untampered regions by leveraging differences in feature distribution without relying on specific forgery indicators.
    \item ECNet \cite{hao2024ec}: EC-Net is an advanced image tampering localization network that employs a two-phase strategy incorporating edge distribution guidance and contrastive learning to accurately identify and segment manipulated regions in images.
\end{itemize}
\begin{table*}[tb]
    \centering
    \renewcommand{\arraystretch}{1.2}
    \small
    \setlength{\tabcolsep}{1pt}
    \begin{tabularx}{\linewidth}{>{\arraybackslash}p{3.8cm} *{4}{>{\centering\arraybackslash}X}}
        \Xhline{2\arrayrulewidth}
        \textbf{Method} & \textbf{Parameters (M)} & \textbf{GFLOPs} & \textbf{Training time (mins)} & \textbf{Inference time (ms)} \\
        \hline
        InpDiffusion & 105.46 & 76.75 & 7.13 & 27.93 \\
        MVSS-Net & 156.92 & 56.56 & 5.75 & 18.83 \\
        EC-Net & 86.52 & 79.28 & 7.91 & 24.76 \\
        \Xhline{2\arrayrulewidth}
    \end{tabularx}
    \normalsize
    \caption{Comparative metrics for different methods including parameters, GFLOPs, training time, and inference time.}
    \label{tab:comparison-metrics}
\end{table*}

\begin{figure}[tb]
    \centering
    \includegraphics[width=1\linewidth]{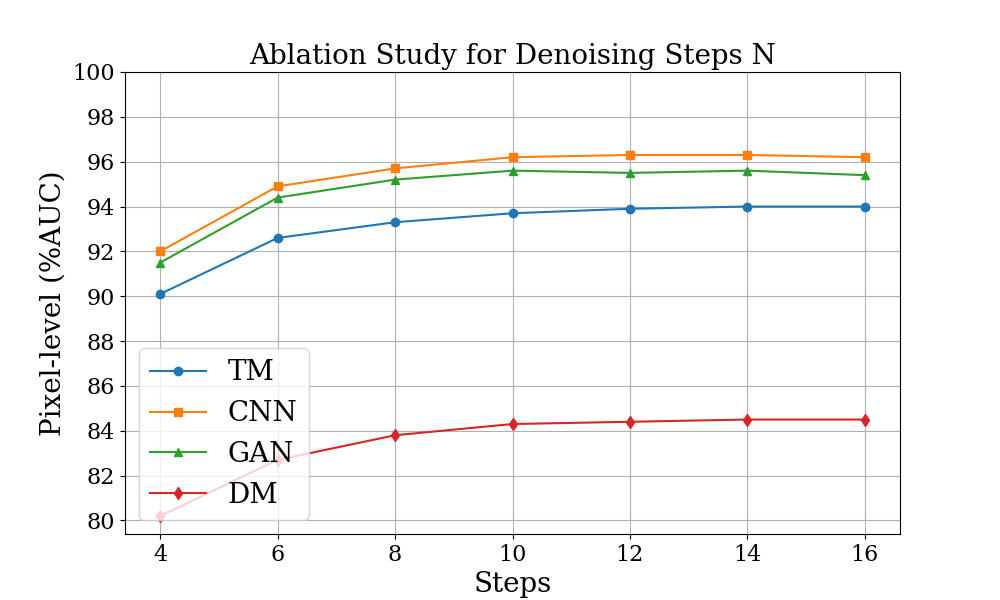}
    \caption{Performance on different denoising steps N.}
    \label{fig:auc_vs_step}
\end{figure}
\subsection{Datasets Details}
The datasets used in this paper are described in detail as follows:
\begin{itemize}
    \item  Inpaint32K \cite{Inpaint32K} :The Inpaint32K dataset is a high-quality collection designed for image inpainting tasks, featuring a diverse range of image sources and scenarios to ensure comprehensiveness and authenticity. It encompasses 32,000 tampered images across four inpainting techniques, including traditional methods, CNN-based, GAN-based, and DM-based approaches. Each inpainting technique consists of 8,000 inpainted images. The dataset includes three types of tampering: replacement, filling, and removal, with each image meticulously crafted to reflect real-world challenges. The dataset is publicly accessible to support research and evaluation in image inpainting localization.
    \item  DID \cite{wu2021iid}: DID dataset includes ten representative inpainting methods, six of which are deep learning-based ncluding GC \cite{GC}, CA \cite{CA}, SH \cite{SH}, EC \cite{EC}, LB \cite{LB} and RN \cite{RN}, and four traditional methods, including NS \cite{NS}, LR \cite{LR}, PM \cite{PM} and SG \cite{SG}. Each of these ten inpainting methods contributed 1000 inpainted images. 
    \item Nist \cite{nist_nimble_2016}: Nist contains 584 image samples with ground-truth masks. Samples from NIST are manipulated using splicing, copy-move, and removal techniques, and are post-processed to hide visible traces.
    \item AutoSplice \cite{jia2023autosplice}: AutoSplice is an AIGC dataset constructed using the DALL-E2 model for automatic image editing. It includes a total of 3,621 images.
    \item IMD \cite{novozamsky2020imd2020}: IMD is a real-life manipulation dataset created by unknown individuals and collected from the Internet. As a result, it includes a variety of manipulation types. The dataset comprises a total of 2,010 image samples.

\end{itemize}

\subsection{Ablation of Hyperparameters}
\Cref{fig:auc_vs_step} presents the performance of InpDiffusion across different sampling steps on images with various types of inpainting techniques. The results indicate that as the number of sampling steps increases, the model's performance improves. However, after exceeding 10 sampling steps, there is no significant improvement in performance. Ultimately, we set the number of sampling steps to 10 to balance performance and computational cost.

Balancing mask supervision and edge supervision in InpDiffusion is crucial for accurately locating locally inpainted objects. \Cref{fig:lambda_mu_ratio_vs_auc} illustrates the detection performance of InpDiffusion under different loss ratios between the two types of supervision. When the loss ratio of mask supervision to edge supervision is 7:3, InpDiffusion achieves the highest performance across all types of inpainted image detections.

\begin{figure}[tb]
    \centering
    \includegraphics[width=1\linewidth]{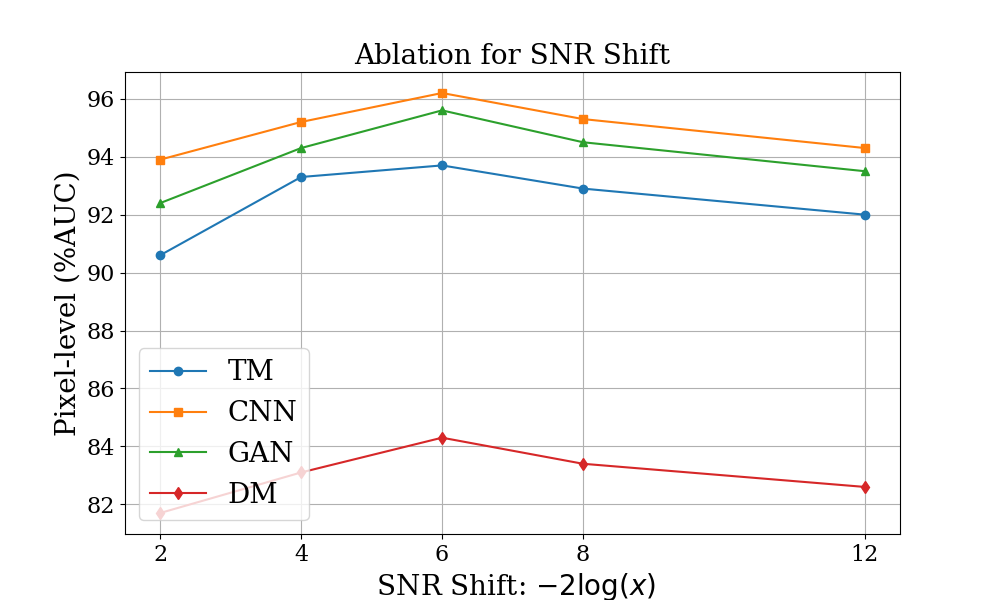}
    \caption{Performance on different SNR Shift during the training phase.}
    \label{fig:snr_shift_vs_auc}
\end{figure}

\begin{figure}[tb]
    \centering
    \includegraphics[width=1\linewidth]{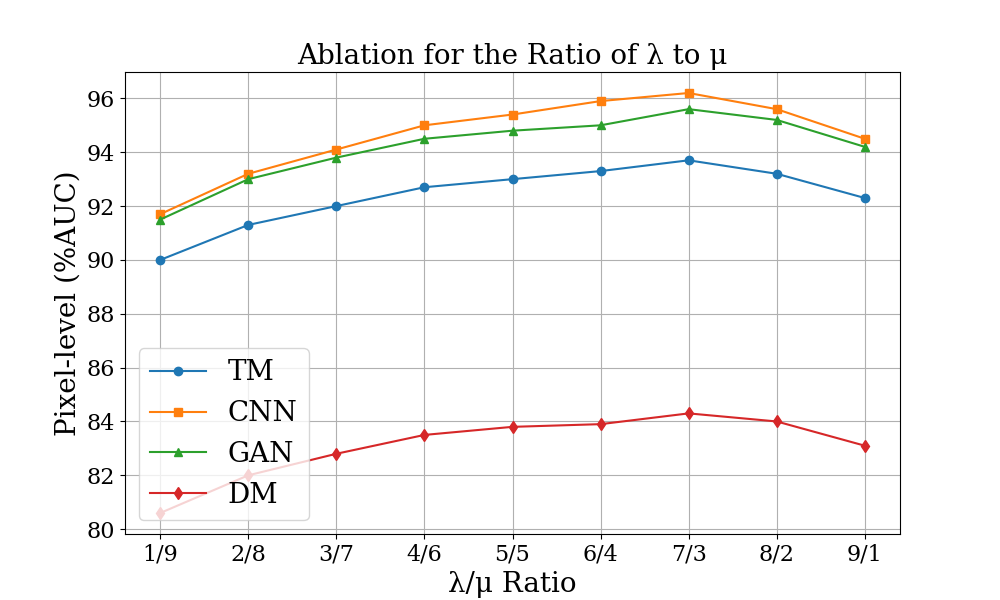}
    \caption{Performance on different ratio of $\lambda$ to $\mu$ in the loss function of InpDiffusion .}
    \label{fig:lambda_mu_ratio_vs_auc}
\end{figure}

Moreover, the impact of SNR Shift on model performance is crucial. As presented in \Cref{fig:SNR shift}, initially setting the SNR shift to $-2\log(12)$ allows the noisy mask to remain distinguishable even at $t=500$ timestamps, suggesting a high signal-to-noise ratio. Consequently, this may lead to an abundance of straightforward training examples for InpDiffusion, potentially resulting in a suboptimal model. In contrast, when the SNR shift is changed to $-2\log(6)$, the noisy mask becomes challenging to observe at $t=500$, providing the model with more challenging and useful information. More quantitative experimental results for selecting SNR Shift are shown in \Cref{fig:snr_shift_vs_auc}.


\subsection{Computational Complexity Analyses}
The proposed InpDiffusion is compared with popular baselines (MVSS-Net and EC-Net) which have available source code for computational complexity analysis, and the results are presented in \Cref{tab:comparison-metrics}. All methods utilize 256×256 forged images as input on a single NVIDIA A800 GPU. We will present a comprehensive evaluation that includes the total training time over 10 epochs on the NIST dataset and the average inference time calculated from a random selection of 500 samples. 
As evidenced by our experiments, the complexity of InpDiffusion is comparable to SOTA methods, and in some aspects, it is even more efficient. While the inference time of InpDiffusion is slightly higher than those of MVSS-Net and EC-Net due to the inherent time metrics associated with the denoising network in diffusion models, the significant improvement in the localization capability for tampered images justifies this trade-off.

\end{document}